\documentclass[journal,transmag]{IEEEtran}
\usepackage{graphicx}
\usepackage{graphics}
\usepackage{subfigure}
\usepackage{calligra}
\ifCLASSINFOpdf
\else
\fi
\begin{document}
%
\title{Demystifying AlphaGo Zero as AlphaGo GAN}



%
%

\author{\IEEEauthorblockN{Xiao Dong, Jiasong Wu, Ling Zhou}
\IEEEauthorblockA{Faculty of Computer Science and Engineering, Southeast University, Nanjing, China}}

%



\IEEEtitleabstractindextext{%
\begin{abstract}
The astonishing success of AlphaGo Zero\cite{Silver_AlphaGo} invokes a worldwide discussion of the future of our human society with a mixed mood of hope, anxiousness, excitement and fear.
We try to dymystify AlphaGo Zero by a qualitative analysis to indicate that AlphaGo Zero can be understood as a specially structured GAN system which is expected to possess an inherent good convergence property. Thus we deduct the success of AlphaGo Zero may not be a sign of a new generation of AI.
\end{abstract}

}

\maketitle



\IEEEdisplaynontitleabstractindextext

%
\IEEEpeerreviewmaketitle

\section{Motivation}
The success of AlphaGo Zero  is definitely a milestone of AI technology. Here we are not trying to discuss the social, cultural or even the ethic impact of it. Instead we are interested in the following technical question: Why can AlphaGo Zero achieve its convergency with a limited self-play generated sampling data and a limited computational cost?

The factors that may influence the performance of AlphaGo Zero include (a)the inherent property of the game Go and (b) the structure of AlphaGo Zero (the ResNet based value and policy network, MCTS and the reinforcement learning structure).

In this paper we try to give a qualitative answer to this question by indicating that AlphaGo Zero can be understood as a special GAN with an expected good convergence property. In another word, the research of GAN gives conditions on the convergency of GANs and AlphaGo Zero seems to fulfill them well.

\section{GAN and its convergence}
Generative Adversarial Networks (GANs)\cite{Wasserstein_GAN} are designed to approximate a distribution $p(x)$ given samples drawn from $p(x)$ with two competing models fighting against each other: a generative model $G$ that capture the data distribution and a discriminative model $D$ that distinguishes the training samples and generated \emph{fake} data samples.

It's well known that the training of GANs is more difficult than normal deep convolutional neural networks (CNNs) due to the following aspects:
\begin{itemize}
  \item In GANs there are two networks, the generator $G$ and the discriminator $D$, need to be trained so that we have to deal with a higher complexity.
  \item There may exist a mismatch between the discriminator $D$ and the generator $G$, which leads to a mode collapse problem.
  \item The cost function may fail to capture the difference between the distributions of the training samples and the generated data as discussed in \cite{}.
\end{itemize}

To deal with the above mentioned problems, various solutions were proposed to improve the convergence of GANs. The main ideas include

\begin{itemize}
  \item \textbf{To eliminate the complexity:}
   This is achieved by adding constraints to the network structureExamples of this strategy include infoGAN and LAPGAN, which reduce the complexity of the generator $G$ by introducing constraints on clusters or subspaces of generated data.
  \item \textbf{To improve the cost function:}
   The most successful example of this class is WGAN which proposed the Wasserstein distance so that the convergence of GANs is almost solved. With Wasserstein distance, the difference of the distributions of the training data and generated data can be reliably captured and the mismatch between the training of $D$ and $G$ is not a serious problem any more.
\end{itemize}

\subsection{The geometry of GAN}

In order to analysis AlphaGo Zero, here we introduce a geometrical picture of GANs\cite{Dong_geometry}, which provides an intuitive understanding of GANs.

In the language of the geometry of deep learning, CNNs and ResNets are all curves in the space of transformations. Accordingly GANs are represented as a two-segment curve since there are two networks, $G$ and $D$ in GANs. From the geometric point of view, the reasons that GANs are difficult to be trained can be understood geometrically as follows:
\begin{itemize}
  \item The higher system complexity of GANs lies in a larger length of the two-segment curve to be found by the training process.
  \item The essential goal of GANs is to train the generative model $G$ by the training data. This means that it's preferred that the information from the training data can directly be fed to $G$. Instead in GANs the information flow from the training data samples to $G$ has to firstly pass the discriminator $D$. Obviously the training of $G$ is highly dependent on $D$ so that usually a balance between the training of $D$ and $G$ need to be carefully adjusted.
  \item The longer information flow path also leads to a serious information loss. An example is that before the Wasserstein distance was introduced, the difference between the distributions of the training data and the generated data can be easily lost so that a failure of convergence may happen.
  \item From the intuitive geometrical picture, training GANs is to find a two-segment curve connecting the input space of $G$ and the decision output space of $D$ while keeping the curve passing a neighbourhood of the training data space in an elegant way, i.e. no mode collapse. But the information flow passway shows that we can not directly see if the curve passes the neighbourhood of the training samples. Instead we can only make an evaluation at the end point of the curve, the output of $D$.
\end{itemize}

Besides these, since GANs are usually based on CNN or ResNets, GANs will also befinite from the strategies to improve the convergence performance on CNN and ResNets. For example from the geometrical point of view, the spectral normalization on GANs \cite{Nobody_SpectralN} can be understood as to set constraints on the Remannian metric of the transformation manifold to control the curvature so that the geodesic shooting like GSD will be more stable. For more details on the geometric picture of deep learning, please refer to \cite{}. In this paper we will only focus on the structure of AlphaGo Zero.

\begin{figure}
  \centering
  \includegraphics[width=7.5cm]{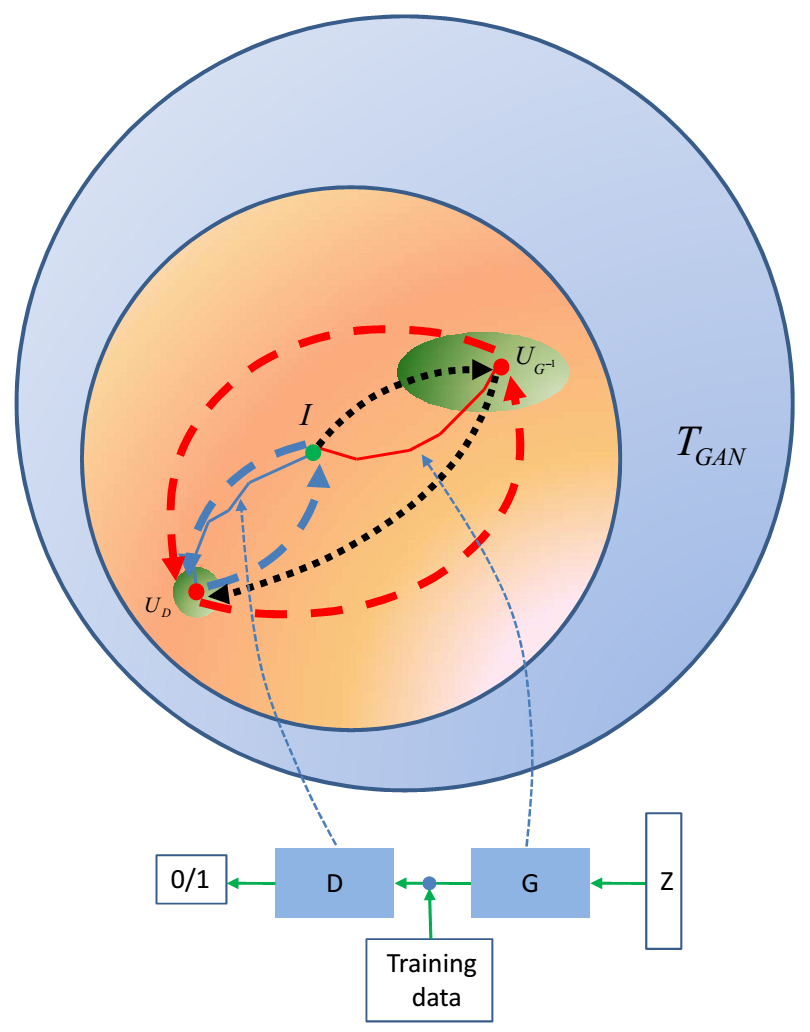}
  \caption{The geometry of GANs. The generator $G$ and discriminator $D$ of a GAN correspond to two curves connecting the identity operation $I$ with the transformations $U_{D}$ and $U_{G^{-1}}$ respectively. During the training procedure, the information flow (the feed-forward and the back-propagation of information) of D and G are shown by the blue and red dash lines. In the design of the GANs, the information flow is shown in the black dash line. The unsupervised learning of the generator G of GANs leads to a much flexible target transformation $U_{G^{-1}}$ compared with the target transformation $U_{CNN}$ in CNNs. }\label{fig-GAN}
\end{figure}

Accordingly, to improve the performance of GANs, we can
\begin{itemize}
  \item Reduce the complexity of GANs by setting constraints on the structure of the networks, or equivalently to reduce the possible \emph{shapes} of the curves.
  \item Directly feed the information from the training data to $G$ so that the information loss problem can be improved.
  \item Find a way to balance the training of $D$ and $G$ to avoid the information loss and mode collapse problems.
\end{itemize}

In the next section, we will show AlphaGo Zero can be understood as a specially designed GAN, whose structure naturally fulfill the above mentioned conditions. We claim that this is the reason that AlphaGo Zero shows a excellent convergency property from the structural point of view.

\section{AlphaGo Zero as GAN}

According to \cite{Silver_AlphaGo}, AlphaGo Zero combines the original value and policy network into a single ResNet to compute the value and policy for any stage $s$ of the game as $(\textbf{P}(s),V(s))=f_{\theta}(s)$. The neural network $f_{\theta}$ is trained by a self-play reinforcement learning algorithm using MCTS to play each move. In each position $s$, an MCTS search guided by $f_{\theta}$ working as a \emph{policy improvement operator} is executed to generate a stronger move policy $\pi$. The self-play using the improved policy $\pi$ to select each move and the game winner $z$ is regarded as a sample of the value, or a \emph{policy evaluation operator}. The reinformcement learning algorithm uses these operators repeatedly to improve the policy, which is achieved by updating the network parameters to make the value and policy $(\textbf{P}(s),V(s))=f_{\theta}$ match the improved policy and self-play winner $(\pi,z)$. The updated network parameters are used for the next iteration to make the search stronger till converge.

Now we can describe AlphaGo Zero in the language of GANs as follows.
\begin{itemize}
  \item The discriminator $D$ is a cascaded series of the same network $f_{\theta}$ connecting from the first move to the end of the game.
  \item The generator $G$ is a cascaded series of the MCTS improved policy guided by $f_{\theta}$, which generates the self-play data.
  \item From a graphical model point of view, the MCTS enhanced policy can be roughly understood as the result of a nonparametric belief propagation(NBP) on a tree.
\end{itemize}

So we can establish a GAN structure on AlphaGo Zero system. We call it AlphaGo GAN since the other versions of AlphaGo can also be regarded as GANs with a minor modification. It should be noticed that in this GAN, both the training data and generative data are generated by the generator $G$ during the self-play procedure.

\subsection{Demystify AlphaGo Zero as GAN}

We can now check if the AlphaGo GANs fulfill the conditions for a good convergence performance of GANs.
\begin{itemize}
  \item \textbf{Complexity:} Both the discriminator $D$ and the generator $G$ hold a similar repeated structure. So although the lengthes of $D$ and $G$ are huge but the complexity is restricted roughly by the size of the policy and value network $f_{\theta}$.
  \item \textbf{Information flow and information loss:} In general GANs, the information of training data can only be fed to $G$ through the output of $D$. Or $G$ is trained by the output of $D$. But in AlphaGo GANs, $G$ is not updated by data based training. Instead it's directly updated by running a MCTS or a NBP based on the information from $D$. We note that the update of $G$ not only depends on the final output of $D$, instead it includes the intermediate information of $D$, i.e. the output of $f_{\theta}$ at every move of the game. Obviously the information of the self-play generated training data and the information of $D$ can be fed to $G$ efficiently.
  \item \textbf{Mismatch between $D$ and $G$:} It was indicated that the mismatch between $D$ and $G$ will lead to either a slow converge or a mode collapse. In AlphaGo GANs, $G$ is a NBP enhanced version of $D$ so that the matching between $D$ and $G$ can be guaranteed.
   \item \textbf{Training data and generated adversarial data:}  In the two-segment curve picture of GANs, we ask that the generated data should pass the neighbourhood of training data. In a general GAN, this can only be justified by checking the outputs of $D$ on the training data and generated data. This is to say, only by checking the distributions of the ouputs of the discriminator $D$ of the training data and the generated data, we can judge if the training data and generated data have the same distribution. In AlphaGo Zero, all the data are generated from self-play using the same policy. So naturally the generated data fall in the neighbourhood of the training data. In another word, the winner and loser's moves are based on the same knowledge and they are just samples from the same distribution.
\end{itemize}

So we can easily see that AlphaGo GANs fulfill the conditions of a \emph{good} GAN. It's not suprising that AlphaGo Zero show a good convergence property.

Based on the GAN structure of AlphaGo Zero, we can then explain the following observations on AlphaGo Zero.

\begin{itemize}
  \item \textbf{Why AlphaGo Zero converges:}
  The good AlphaGo GAN structure is only one reason for the convergence of AlphaGo Zero. We have to assume that the problem itself, the game of Go, should hold an elegant structure such that the convergence can be achieved. This may be a hint that the successness of AlphaGo Zero may not be regarded as a universal phenomenon since the convergence is highly dependent on the problem itself.

  \item \textbf{Why human knowledge deteriorates its performance:}
  It's observed that a pre-training using human knowledge can result in a worse performance. In GAN's language, the pre-training will lead to human knowledge biased strong policy, or an over-strong discriminator. It's well known that a over-strong discriminator will lead to a deteriorated convergence. Or in the language of NBP, the discriminator is so strong that the NBP based generator $G$ can not further enhance or shift it.
\end{itemize}

\section{Conclusions}

In this work we try to understand AlphaGo Zero as a GAN structure called AlphaGo GAN. Combining the geometrical picture of deep learning, we show that AlphaGo Zero can be analyzed as a GAN with a special structure, which fulfills the good convergence conditions of GANs. We then conclude that convergence of AlphaGo Zero is a joint result of both the special structure of the game GO and the structure of AlphaGo GAN. The success of AlphaGo Zero is not mysterious and it's not safe to claim that this can be generalized to other applications.

\bibliographystyle{unsrt}

\bibliography{AlphaGo}





\end{document}